\documentclass{article}
\usepackage{spconf,amsmath,graphicx}

\usepackage{multirow}
\usepackage{amsmath}
\usepackage{amssymb}
\usepackage{url}

\title{Scalable Neural Dialogue State Tracking}

\name{Vevake Balaraman$^{1,2}$ and Bernardo Magnini$^{1}$}
\address{
  $^1$Fondazione Bruno Kessler, Trento, Italy \\
  $^2$ICT Doctoral School, University of Trento, Italy\\
  \textit{\{balaraman, magnini\}@fbk.eu}}

\begin{document}

\maketitle

\begin{abstract}

A Dialogue State Tracker (DST) is a key component in a dialogue system aiming at estimating the beliefs of possible user goals at each dialogue turn.
Most of the current DST trackers make use of recurrent neural networks and are based on complex architectures that manage several aspects of a dialogue, including the user utterance, the system actions, and the slot-value pairs defined in a domain ontology. However, the complexity of such neural architectures incurs into a considerable latency in the dialogue state prediction, which limits the deployments of the models in real-world applications, particularly when task scalability (i.e. amount of slots) is a crucial factor.
In this paper, we propose an innovative neural model for dialogue state tracking, named \textbf{G}lobal encoder and \textbf{S}lot-\textbf{At}tentive decoders (G-SAT), which can predict the dialogue state with a very low latency time, while maintaining high-level performance. 
We report experiments on three different languages (English, Italian, and German) of the WOZ2.0 dataset, and show that the proposed approach provides competitive advantages over state-of-art DST systems, both in terms of accuracy and in terms of time complexity for predictions, being over 15 times faster than the other systems.

\end{abstract}
\begin{keywords}
Dialogue state tracking, deep learning, dialogue systems
\end{keywords}

\section{Introduction}
\label{sec:intro}
\begin{figure}
    \centering
    \includegraphics[width=0.9\columnwidth]{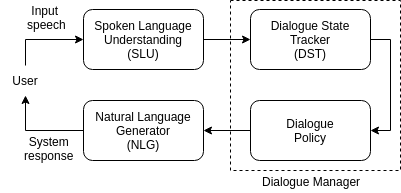}
    \caption{A typical flow in a task-oriented dialog system \cite{Young2013}.}
    \label{fig:system}
\end{figure}

Spoken dialogue systems, or conversational systems, are designed to interact and assist users using speech and natural language to achieve a certain goal \cite{Henderson2015}.
A consolidated approach to build a task-oriented dialogue system involves a pipeline architecture (see Figure \ref{fig:system}), where each component is trained to perform a sub-task, and the combination of the modules in a given sequence aims at handling the complete task-oriented dialogue.  In such a pipeline, a spoken language understanding (SLU) module determines the user's intent and the relevant information that the user is providing represented in terms of slot-value pairs. Then, the dialogue state tracker (DST) uses the information of the SLU together with the past dialogue context and updates its belief state \cite{Wang2013, Sun2015a}. In this framework a dialogue state indicates what the user requires at any point in the dialogue, and it is represented as a probability distribution over the possible states (typically a set of pre-defined slot-value pairs specific of the task). The dialogue policy manager, then, decides on the next system action based on the dialogue state. Finally, a natural language generation (NLG) component is responsible for the generation of an utterance that is returned as response to the user.

In this paper we focus on the dialogue state tracker component, whose role is to track the state of the dialogue based on the current user utterance, the conversation history and any additional information available to the system \cite{Henderson2015}. 
Deep neural network techniques, such as recurrent neural network and convolutional neural networks, are the current state-of-the-art models for DST \cite{Henderson2014, Wen2017, NBT, Ren2018, GLAD}, showing high capacity to generalize from annotated dialogues. As an example, the GLAD model (Global-Locally Self-Attentive Dialogue State Tracker - \cite{GLAD}),  employs independent slot-specific encoders, consisting of a recurrent and a self-attention layer, for each of the user utterance, the system action and the slot-value pairs. Another DST system, GCE (Globally Conditioned Encoder - \cite{GCE}), simplifies the GLAD neural architecture removing the slot-specific recurrent and self-attention layers of the encoder, but still requires separate encoders for the utterance, the system action and the slot-values.

Although the neural network models mentioned above achieve state-of-the-art performance, the complexity of their architectures make them highly inefficient in terms of \textit{time complexity}, with a significant latency in their prediction time. Such latency may soon become a serious limitation for their deployment into concrete application scenarios with increasing number of slots, where real time is a strong requirement. Along this perspective, this work investigates the time complexity of state-of-the-art DST models and addresses their current limitations. Our contributions are the following:
\begin{itemize}
    \item we have designed and implemented an efficient DST, consisting of a \textbf{G}lobal encoder and \textbf{S}lot-\textbf{At}tentive decoders (G-SAT);
    \item we provide empirical evidences (three languages of the WOZ2.0 dataset \cite{NBT}) that the proposed G-SAT model considerably reduces the latency time with respect to state-of-art DST systems (i.e. over 15 times faster), while keeping the dialogue state prediction inline with such systems;
    \item further experiments show that the proposed model is highly robust when either  pre-trained embeddings are used or when they are not used, in this case outperforming state-of-art systems. 
\end{itemize}

The implementation of the proposed G-SAT model is publicly available\footnote{\url{https://github.com/vevake/GSAT}}.

The paper is structured as follows. Section 2 summarizes the main concepts behind the definition of dialogue state tracking. Section 3 reports the relevant related work. Section 4 provides the details of the proposed G-SAT neural model, and, finally, Section 5 and 6 focus on the experiments and the discussion of the results we achieved.

\section{Dialogue State Tracking}

\begin{figure}
    \centering
    \includegraphics[width=\columnwidth]{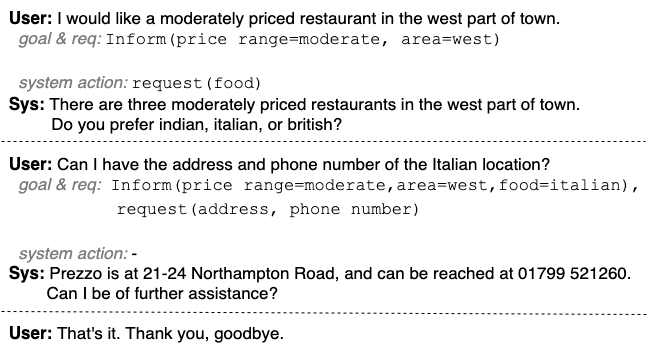}
    \caption{An annotated dialogue from the WOZ2.0 dataset, with each turn separated by a dashed line.}
    \label{fig:example}
\end{figure}

A Dialogue State Tracker (DST) estimates the distribution over the values $V_s$ for each slot $s \in S$ based on the user utterance and the dialogue history at each turn in the dialogue.
Slots $S$ are typically provided by a domain ontology, and they can either be \textit{informable} ($S_{inf}$) or \textit{requestable} ($S_{req}$).
\textit{Informable} slots are attributes that can be provided by the user during the dialogue as constraints, while  \textit{requestable} slots are attributes that the user may request from the system.
The dialogue state typically maintains two internal properties: 
\begin{itemize}
    \item \textit{joint goal} - indicating a value $v \in V_s$ that the user specifies for each informable slot $s \in S_{inf}$.
    \item \textit{requests} - indicating the information that the user has asked the system from the set of requestable slots $S_{req}$.
\end{itemize}
For example, in the restaurant booking dialogue shown in Figure \ref{fig:example}, extracted from the the WOZ2.0 dataset \cite{NBT}, the user  specifies a constraint on the \textit{price range} slot (i.e. \textit{inform(price range=moderate)}) in the first utterance of the dialogue, and requests the phone number and the address (i.e. \textit{request(address, phone number)}) in the second user utterance.
The set of slot-value pairs specifying the constraints at any point in the dialogue (e.g. (\textit{price\ range=moderate, area=west, food=italian})) is referred to as the \textit{joint goal}, while the requested slot-value pairs for a given turn (e.g. \textit{request(address, phone number)}) as \textit{turn request}.

\subsection{Latency in Dialogue Systems}
An effective dialogue system should be able to process the user utterance and respond in real-time in order to achieve a smooth dialogue interaction between the user and the dialogue system itself \cite{nakano1999, challenges}. As a consequence, time latency in dialogue systems is very crucial, as it directly impacts the user experience.
In real world applications, task-oriented dialogue systems typically follow a pipeline architecture (as shown in Figure \ref{fig:system})
where multiple components need to interact with each other to produce a response for the user, the time complexity of each component plays a key role.
In particular, the DST component is a bottleneck, as it needs to track the user's goals based on the dialogue and provides the output to other components of the whole process.
Though end-to-end (E2E) approaches for dialogue systems have attracted recent research, dialog state tracking still remains an integral part in those systems, as shown by \cite{Liu2017, Wen2017, Li2017}.

Current DST models use recurrent neural networks (RNN), as they are able to capture temporal dependencies in the input sentence.
A RNN processes each token in the input sequentially, one after the other, and so can incur significant latency if not modeled well.
Apart from the architecture, the number of slots and values of the domain ontology also affects the time complexity of the DST.
Recent works \cite{NBT, GLAD, Ren2018} use RNNs to obtain very high performance for DST, but nevertheless are quite limited as far as the efficiency of the models are concerned.
For instance, the GCE model \cite{GCE} addresses time complexity within the same architectural framework used by of GLAD \cite{GLAD}, although the latency prediction of the model is still quite poor, at least for a production system (more details in Section \ref{sec:exp}).
This limitation could be attributed to the fact that both GLAD and GCE use separate recurrent modules to output representations for user utterance, system action and slot-value pairs.
These output representations need then to be combined using a scoring module which scores a given slot-value pair based on the user utterance and the system action separately.
In this work, we investigate approaches that overcome the complexity of such architectures and improve the latency time without compromising the DST performance.

\section{Related Work}
\label{sec:rel_work}
Spoken dialogue systems typically consist of a spoken language understanding (SLU) component that performs slot-filling to detect slot-value pairs expressed in the user utterance.
This information is then used by dialogue state tracker(DST) \cite{Wang2013, Sun2015a}.
Recent research has focused on jointly modeling the SLU and the DST \cite{Henderson2014, Wen2017}.
For such joint models, deep neural network techniques have been the choice of use because of their proven ability to extract features from a given input and their generalization capability \cite{Henderson2014, Wen2017, NBT, Ren2018}.
Following this research line, \cite{Henderson2014} proposed a word-based DST (based on a delexicalisation approach) that jointly models SLU and DST, and directly maps from the utterances to an updated belief state.
\cite{NBT} proposed a data-driven approach for DST, named neural belied tracker (NBT), which learns a vector representation for each slot-value pair and compares them with the vector representation of the user utterance to predict if the user has expressed the corresponding slot-value pair.
The NBT model uses pre-trained semantic embeddings to train a model without semantic lexicon.

GLAD (Global-Locally Self-Attentive Dialogue State Tracker) \cite{GLAD} consists of a shared global bidirectional-LSTM \cite{hochreiter1997long} for all slots, and a local bidirectional-LSTM for each slot.
The global and local representations are then combined using attention, which then is used by a scoring module to obtain scores for each slot-value pair.
GLAD also relies on pre-trained embeddings, and since it consists of multiple recurrent modules, the latency of the model is quite high.
In \cite{GCE}, a Globally conditioned encoder (GCE) is used as a shared encoder for all slots and aim to address this issue by proposing a single encoder with global conditioning.
While this approach reduced the latency of GLAD, it still has a considerable time complexity for real-world applications, which is discussed in Section \ref{sec:results}.

StateNet, proposed by \cite{Ren2018}, uses a LSTM network to create a vector representation of the user utterance, which is then compared against the vector representation of the slot-value candidates.
\cite{Ren2018} is the current state-of-art for DST: however it can be used for domains iff pre-trained embeddings exist and can only be modelled for informable slot and not for the requestable slots.
\cite{Mandy2018} used convolutional neural network (CNN) for DST and showed that without pre-trained embeddings or semantic dictionaries, the model can be competitive to state-of-the-art.

\section{The G-SAT Model}
\label{sec:model}
The proposed approach, G-SAT (\textbf{G}lobal encoder and \textbf{S}lot-\textbf{At}tentive decoders), is designed to predict slot value pairs for a given turn in the dialogue.
For a dialogue turn, given the user utterance $U$, the previous system action $A$ and the value set $V_s$ for slot $s \in S$, the proposed model provides a probability distribution over slot-value set $V_s$.
\begin{equation}
    P_s = DST(A, U, V_s)
\end{equation}

The model consists of a single encoder module and a number of slot specific decoder (classifier) modules.
The encoder consists of a recurrent neural network that takes as input both the user utterance $U$ and the previous system action $A$, and outputs a vector representation $h$.
The classifier then receives the representation $h$ and the slot-values $v \in V_s$ and estimates the probability of each value in a given slot.

\subsection{Encoder}
The encoder takes in the previous system action $A$ and the current user utterance $U$ as inputs, and processes them iteratively to output a hidden representation for each token in the input, as well as the context vector.

Let the user utterance at time $t$ be denoted as $U=\{u_1,u_2,...,u_k\}$ with $k$ words and $A$ denotes the previous system action.
The system action $A$ is converted into a sequence of tokens that include the $action, slot$ and $value$ (e.g. \textit{confirm(food=Italian)} $\rightarrow$ confirm food Italian) and is denoted as $A=\{a_1, a_2,..,a_l\}$.
In case of multiple actions expressed by the system, we concatenate them together.
The user utterance $U$ and system action $A$ are then concatenated forming the input $X$ to the encoder.
\[X = [a_1,...a_l;u_1,..u_k] = [x_1,x_2,...x_n]\]
where $[\:;\:]$ denotes concatenation.
Each input tokens in $\{x_1,x_2,..x_n\}$ is then represented as a vector 
$\{\boldsymbol{x_1},\boldsymbol{x_2},..,\boldsymbol{x_n}\}$ 
by an embedding matrix $E \in \mathbb{R}^{|v|\times d}$,
where $|v|$ is the vocabulary size and $d$ is the embedding dimension.
This representation is then input to a bidirectional-LSTM \cite{hochreiter1997long} that processes the input in both forward and backward directions, to yield the hidden representations, as follows:
\begin{align}
    \overrightarrow{h}_t &= LSTM_f(\overrightarrow{h}_{t-1}, \boldsymbol{x_t}) \\
    \overleftarrow{h}_t &= LSTM_b(\overleftarrow{h}_{t+1}, \boldsymbol{x_t})
\end{align}
where $LSTM_f(.)$ and $LSTM_b(.)$ are the forward and backward LSTMs. $\overrightarrow{h}_t$ and $\overleftarrow{h}_t$ are the corresponding hidden states of forward and backward LSTMs at time $t$.
The representations $h_t$ for each token in the input and the overall input representation $h_L$ are then obtained as follows:
\begin{align}
    h_t &= [\overrightarrow{h}_t;\overleftarrow{h}_t] \\
    h_L & = [\overrightarrow{h}_n;\overleftarrow{h}_1]
\end{align}

Since our model uses a shared encoder for all slots, the outputs of the encoder $h_t$ and $h_L$ are used by slot specific classifiers for prediction on corresponding slots.

\subsection{Classifier}
The classifier predicts the probability for each of the possible values $v \in V_s$  of a given slot $s \in S$.
It takes in the hidden representations $h_t$ and $h_L$ of the encoder, and the set of possible values $V_s$ for a given slot $s$, and computes the probability for each value being expressed as a constraint by the user.
Initially, each of the slot-value is represented by a vector $\boldsymbol{v}$, using the same embedding matrix $E$ of the encoder.
For slot-values with multiple tokens, their corresponding embeddings are summed together to yield a single vector.
The embeddings are then transformed as following to obtain a representation of the slot values:
\begin{equation}
    Z_s = W_s\boldsymbol{V}_s^T \\
\end{equation}
where $\boldsymbol{V}_s = \{\boldsymbol{v_1}, \boldsymbol{v_2},..\}$ for slot $s$, and $W_s$ is the parameter learned during training.
The encoder hidden state $h_L$ is then transformed using the $ReLU$ activation function, to obtain a slot specific representation of the user utterance, as follows:
\begin{equation}
    U_s = ReLU(W_hh_L)
\end{equation}

Based on the slot specific input representation $U_s$, an attention mechanism weights the hidden states of the input tokens $h_t$ to provide a context vector $C$.
\begin{align}
    a_i &= tanh(W_c[U_s;h_i]) \\
    \alpha &= Softmax(a) \\
    C &= \sum_i \alpha_ih_i
\end{align}

Depending on the slot type (informable or requestable), the final layer of the classifier varies.

\subsubsection{Informable slots}
The context vector $C$ and the slot-value representations $Z_s$ are then used to obtain the probability for each slot-value as follows:
\begin{align}
    score &= C \cdot Z_s \\
    \psi_p^i &= Softmax([score_{none}; score])
\end{align}
where $score_{none}$ is the score for $none$ value, and it is learned by the model. $\psi_p^i$ is the probability of the slot-value pair expressed as constraint.

\subsubsection{Requestable slots}
For requestable slots, we model a binary prediction for each possible requestable slot as follows:
\begin{align}
    score &= C \cdot Z_s \\
    \psi_p^r &= Sigmoid(score)
\end{align}

where $\psi_p^r$ contains the probability for each requestable slot being requested by the user.

\section{Experiments}
\label{sec:exp}
In this section we describe the dataset and the experimental setting used for the dialogue state tracking task.

\subsection{Datasets}
We use the the WOZ2.0 \cite{NBT} dataset, collected using a Wizard of Oz framework, and consisting of written text conversations for the restaurant booking task.
Each turn in a dialogue was contributed by different users, who had to review all previous turns in that dialogue before contributing to the turn.
WOZ2.0 consists of a total of 1200 dialogues, out of which 600 are for training, 200 for development and 400 for testing.
\cite{mrksicsemantic} translated the WOZ.0 English data both to German and Italian using professional translators.
We experiment on the three languages (English, German, Italian) of the WOZ2.0 dataset.

\subsection{Evaluation metrics}
We evaluate our proposed model both in terms of performance and prediction latency.
The performance of the model is evaluated using the standard metrics for dialogue state tracking, namely, \textit{joint goal} and \textit{turn request} \cite{DSTC2}.
\begin{itemize}
    \item \textbf{Joint Goal:} indicates the performance of the model in correctly tracking the goal constraints over a dialogue. The joint goal is the set of accumulated turn level goals up to a given turn.
    \item \textbf{Turn Request:} indicates the performance of the model in correctly identifying the user's request for information at a given turn.
\end{itemize}

The prediction latency of the model is evaluated using \textit{time complexity}.
\begin{itemize}
    \item \textbf{Time complexity:} indicates the latency incurred by the model in making predictions. The time complexity is indicated as the time taken to process a single batch of data.
\end{itemize}

\begin{table}
    \centering
    \begin{tabular}{l|c|c}
        \multicolumn{1}{c|}{\textbf{Model}} & \textbf{Joint Goal} & \textbf{Turn Request}\\
        \hline
        Delexicalisation Model\cite{NBT} & 70.8 & 87.1\\
        NBT - CNN\cite{NBT} & 84.2 & 91.6 \\
        NBT - DNN\cite{NBT} & 84.4 & 91.2 \\    
        CNN\cite{Mandy2018} & 86.9 & 95.4\\
        GLAD\cite{GLAD} & 88.1 & 97.1\\
        GCE\cite{GCE} & 88.5 & 97.4\\
        StateNet\_PSI\cite{Ren2018} & 88.9 & - \\
        \hline
        Our Approach (G-SAT) & 88.7 & 96.9 \\
    \end{tabular}
    \label{tab:exp_results}
    \caption{Dialog state tracking results on the WOZ2.0 English testset.}
\end{table}

\subsection{Experimented Models}
We compared our G-SAT model against seven DST models. The Delexicalisation model \cite{NBT} uses a delexicalisation approach (i.e. replacing slot value tokens with generic terms) and requires large semantic dictionaries, while all the other approaches are data driven.
The neural belief tracker (NBT) \cite{NBT} builds on the advances in representation learning and uses pre-trained embeddings to overcome the requirement of hand-crafted features. 
The convolutional neural network (CNN) model \cite{Mandy2018} is the only approach that does not use pre-trained embeddings, although they use also the development data for model training.
GLAD \cite{GLAD}, GCE \cite{GCE} and StateNet\_PSI \cite{Ren2018} are based on recurrent neural networks
and use pre-trained embeddings.
To facilitate comparison, our G-SAT approach is trained with the same pre-trained embeddings as used in GLAD and GCE.

\subsection{Implementation}
We use the pytorch \cite{Paszke2017AutomaticDI} library to implement our model (G-SAT).
The encoder of the model is shared across all slots and a separate classifier is defined for each slot.
The number of hidden units of the LSTM is set to 64 and a dropout of 0.2 is applied between different layers.
We use Adam optimizer with a learning rate of 0.001.
The embedding dimension of the default model is set to 128, and embeddings are learned during training.
In order to have a fair comparison with other models that use pre-trained embeddings, we also experiment our approach using pre-trained GloVe embeddings (of dimension 300) \cite{glove}, and character n-gram embeddings (of dimension 100) \cite{hashimoto} as used in GLAD, leading to embedding of size 400.
The turn-level predictions are accumulated forward through the dialogue and the goal for slot $s$ is \textit{None} until it is predicted as value $v$ by the model.
The implemented model is experimented with 10 different random initializations for each language, and the scores reported in Section \ref{sec:results} are the mean and standard deviation obtained in the experiments.

\section{Results and Discussion}
\label{sec:results}
In this section we initially discuss  the model performance in terms of \textit{joint goal} and \textit{turn request}; and later we show a comparison of the \textit{time complexity} of the models.

\subsection{DST Performance}
The \textit{joint goal} and \textit{turn request} performance of the experimented models (as they are reported in their respective papers) are shown in Table 1. 
We can see that the G-SAT proposed architecture is comparable with respect to  the other model and outperforms both GLAD and GCE on \textit{joint goal} metric. This shows that G-SAT is highly competitive with the state of the art in DST.

To investigate the behaviour of different models  without any pre-trained embeddings, we use the official implementations of GLAD\footnote{\url{https://github.com/salesforce/glad}} and GCE\footnote{\url{https://github.com/elnaaz/GCE-Model}}, and perform the experiments such that embeddings are learned from the data.
We increased the epochs from 50 (default) to 150 for GLAD and GCE experiments, as we noticed that the model did not converge with 50 epochs (since embeddings are also learned here).
The other parameters of the model are set as the default implementation.
Since the core of the StateNet\_PSI model is to output a representation on which a 2-Norm distance is calculated against the word-vector of the slot-value, which is fixed, it is not suitable to train the embeddings.
For this reason we do not experiment with StateNet\_PSI.
In addition, the StateNet\_PSI model can only predict informable slots (i.e. can not predict requestable slots), unlike the other approaches considered in our experiments.

\begin{table}
    \centering
    \begin{tabular}{c|c|c|c|l}
        \multirow{2}{*}{\textbf{Language}} & \multirow{2}{*}{\textbf{Data}} & \multicolumn{3}{c}{\textbf{Model}}\\
        & & \textbf{GLAD} & \textbf{GCE} & \multicolumn{1}{c}{\textbf{G-SAT}}\\
        \hline
        \multirow{2}{*}{English} &  Dev & 88.4$\pm$0.6 & 89.0$\pm$1.0 & \textbf{89.0}$\pm$0.6 \\
        \cline{2-5}
        & Test & 84.6$\pm$0.7 & 85.1$\pm$1.1 & \textbf{87.6}$\pm$0.6 * \\
        \hline
        \multirow{2}{*}{Italian} &  Dev & 73.2$\pm$1.3 & 72.1$\pm$0.9 & \textbf{76.6}$\pm$1.4 * \\
        \cline{2-5}
        & Test & 76.3$\pm$1.2 & 77.2$\pm$1.7 & \textbf{79.4}$\pm$1.5 * \\
        \hline
        \multirow{2}{*}{German} &  Dev & 52.3$\pm$1.4 & 52.1$\pm$1.0 & \textbf{56.4}$\pm$1.0 * \\
        \cline{2-5}
        & Test & 59.3$\pm$1.9 & 59.8$\pm$1.2 & \textbf{62.4}$\pm$1.4 * \\
        \hline
    \end{tabular}
    \caption{Joint Goal performance on WOZ2.0 test data: all models trained without pre-trained embeddings. * indicates statistically significant \cite{reimers-gurevych-2017-reporting} than both GLAD and GCE, using Wilcoxon signed-rank test (with p$<$0.05).}
    \label{tab:result}
\end{table}

Table \ref{tab:result} shows the joint goal performance of the models on both the development and test data for three different languages.
We can see that our model (G-SAT)  outperforms both GLAD and GCE on the three languages of the WOZ2.0 dataset when no pre-trained resources are available, and that the model performance is consistent across both the development and the test data.

Table \ref{tab:req_result} shows the turn request performance of each model for the three languages.
Even in this case the G-SAT model is very competitive on the three languages compared to both GLAD and GCE models.
In addition, since predicting a requestable slot is a much easier task than predicting  an informable slot, we note that all three models show very high performance.

\begin{table}
    \centering
    \begin{tabular}{c|c|c|c}
        \multirow{2}{*}{\textbf{Language}} & \multicolumn{3}{c}{\textbf{Model}}\\
        & \textbf{GLAD} & \textbf{GCE} &\textbf{G-SAT}\\
        \hline
        English & 97.0$\pm$0.3 & 96.8$\pm$0.3 & \textbf{97.2}$\pm$0.2 \\
        Italian & \textbf{95.9}$\pm$0.2 & \textbf{95.9}$\pm$0.2 & 95.8$\pm$0.2 \\
        German & 94.7$\pm$0.4 & 94.4$\pm$0.4 & \textbf{94.8}$\pm$0.4 \\
    \end{tabular}
    \caption{Turn Request performance on WOZ2.0 test data: all models  trained without pre-trained embeddings.}
    \label{tab:req_result}
\end{table}

\subsection{Time Complexity Performance}
\begin{figure}
    \centering
    \includegraphics[width=\columnwidth]{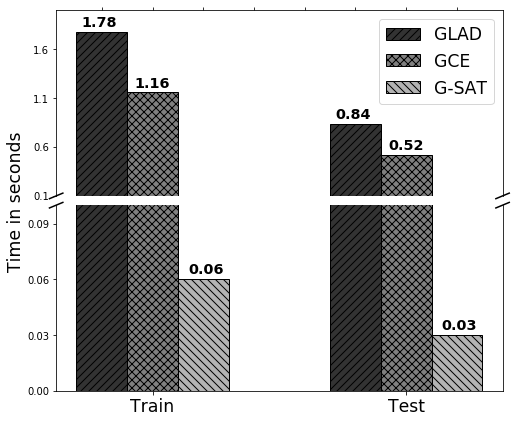}
    \caption{Time complexity of various models for each batch of size 50 during training and testing (low execution time means low latency in prediction).}
    \label{fig:time}
\end{figure}

The time complexity for GLAD, GCE and our model (G-SAT) is shown in Figure \ref{fig:time}.
All models are executed with batch size of 50, under the same environment and hardware (single GeForce GTX 1080Ti GPU).
As the pre-processing and post-processing of each model can vary based on the implementation and the approach, we report only the time complexity for the model execution after it is loaded and ready to be executed.
GLAD requires 1.78 seconds for training for each batch of data, and 0.84 seconds to predict for a batch.
Since  GCE  does not require a separate encoder for each slot, as in GLAD, it reduces the training time to 1.16 seconds, and the prediction time to 0.52 seconds.
Our approach has a significant advantage in the execution time, requiring only 0.06 seconds for training and 0.03 for prediction of each batch.
We notice that, while the time complexity of GLAD and GCE  reported in \cite{GCE} coincide with our results for training,  results on the test data differ considerably. In fact, the time complexity for GCE reported in \cite{GCE} was 1.92 seconds, while in our experiment we found that GCE instead processes 1.92 batches/second leading to 0.52 seconds/batch.

We also considered the number of parameters of the three models, as they have a direct impact on the memory footprint and the training/execution time.
Only the trainable parameters for each model are reported under the default setting.
Since both GLAD and GCE use pre-trained embeddings, the parameter count do not include the embedding size, while our approach includes also the embeddings as parameters as they are learned from the model.
The GLAD model has $\sim$14M trainable parameters, while the GCE model has $\sim$5M parameters.
Since GCE has a single encoder, compared to different encoders for each slot as in GLAD, it reduces the model size to almost one-third.
On the other hand, our G-SAT approach has only $\sim$460K parameters, making it suitable for low memory footprint scenarios. To sum up, GCE  has over 11 times the  parameters than the proposed model, while GLAD has over 31 times the proposed model.

\subsection{Discussion}
Both GLAD and GCE, by default, use  embeddings of size 400, while our G-SAT model has a default embedding size of 128.
So we also investigated the effect of embedding dimension on these different models, to understand if  results are consistent, or if the choice of the embedding size has a significant role in the performance of the models (as the embeddings are learned during training).
First, we experimented our approach with the same embedding size as GLAD and GCE, which is of dimension 400.
In this case G-SAT achieved 88.6 and 86.7 on the dev and test on English, respectively, still outperforming GLAD (dev:88.4, test:84.6) and GCE (dev:89.0, test:85.1).

In a second experiment, we reduced the embedding dimension of both GLAD and GCE to 128, and trained the model.
The performance of GLAD (dev:87.1, test:84.6), GCE (dev:87.8, test:85.6) and G-SAT (dev:89.0, test:87.6) showed again the same trend.

\section{Conclusion}
\label{sec:conclusion}
In this paper we addressed time complexity issues in modelling an effective dialogue state tracker such that it is suitable to be used in real-world applications, particularly where the number of slots for the task becomes very high.
We proposed a neural model, G-SAT, with a simpler architecture compared to other approaches.
We provided experimental evidences that the G-SAT model significantly reduces the prediction time (more than 15 times faster than previous approaches), still performing competitive to the state-of-the-art.
As for future work, we would like to investigate our approach in the case of a multi-domain dialogue state tracking, where the DST should track for multiple domains and the number of slots is much higher compared to single-domain datasets.

\bibliographystyle{IEEEbib}
\bibliography{strings,refs}

\end{document}